\documentclass{article}

\usepackage[final]{nips_2017}
\usepackage[T1]{fontenc}    
\usepackage{hyperref}       
\usepackage{url}            
\usepackage{booktabs}       
\usepackage{amsfonts}       
\usepackage{nicefrac}       
\usepackage{microtype}      
\usepackage{graphicx}
\usepackage{wrapfig}
\usepackage{amsmath}

\title{ObamaNet: Photo-realistic lip-sync from text}

\author{
  Rithesh Kumar\text{*}$^\dagger$, Jose Sotelo\text{*}$^\dagger$, Kundan Kumar\text{*}$^\dagger$, Alexandre de Br\'ebisson\text{*}$^\dagger$, Yoshua Bengio$^\dagger$\\
  \text{*}Lyrebird.ai\\
  $^\dagger$MILA\\
}

\begin{document}

\maketitle

\begin{abstract}

We present \textbf{ObamaNet}, the first architecture that takes any text as input and generates both the corresponding speech and synchronized photo-realistic lip-sync videos. Contrary to other published lip-sync approaches, ours is only composed of fully trainable neural modules and does not rely on any traditional computer graphics methods. More precisely, we use three main modules: a text-to-speech network based on \textbf{Char2Wav}, a time-delayed LSTM to generate mouth-keypoints synced to the audio, and a network based on \textbf{Pix2Pix} to generate the video frames conditioned on the keypoints.

\end{abstract}

\section{Introduction}

There is currently extensive research on machine learning approaches to generate images (\cite{isola2016image}). In parallel, there has been significant progress in speech synthesis (\cite{sotelo2017char2wav}). Nevertheless very little work attempts to model both modalities at the same time. In our work, we show that we can combine some of these recently developed models to generate artificial videos of a person reading aloud an arbitrary text. Our model can be trained on any set of close shot videos of a person speaking, along with the corresponding transcript. The result is a system that generates speech from an arbitrary text and modifies accordingly the mouth area of one existing video so that it looks natural and realistic. A video example can be found there: \mbox{\url{http://ritheshkumar.com/obamanet}}

Although we showcase the method on Barack Obama because his videos are commonly used to benchmark lip-sync methods (see for example \cite{suwajanakorn2017synthesizing}), our approach can be used to generate videos of anyone provided the data is available.

\section{Related Work}

Recently, significant progress in the generation of photo-realistic videos have been made \citep{thies2016face}. In particular \citet{Karras} have tried to generate facial animations based on audio. The work by \cite{suwajanakorn2017synthesizing} is the closest to ours, yet we have two important differences. First, we have a neural network instead of a traditional computer vision model. Second, we add a text-to-speech synthesizer in order to have a full text-to-video system.

\begin{figure}[h!]
    \centering
    \includegraphics[scale=.3]{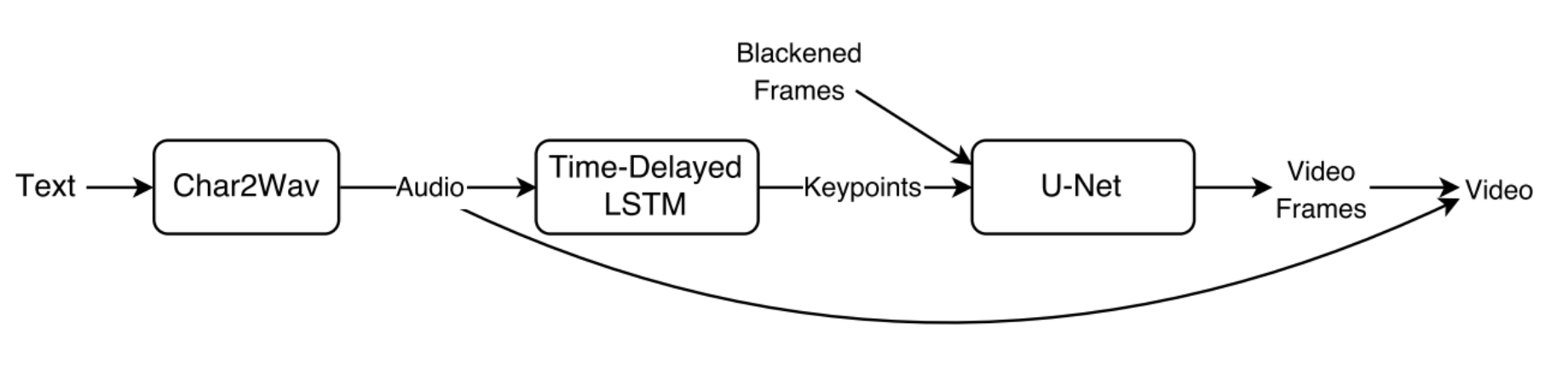}
    \caption{Flow diagram of our generation system.}
    \label{fig:full_system}
\end{figure}

\newpage
\section{Model Description}
\begin{wrapfigure}[22]{l}{0.4\textwidth}
	\begin{center}
        \includegraphics[scale=0.4]{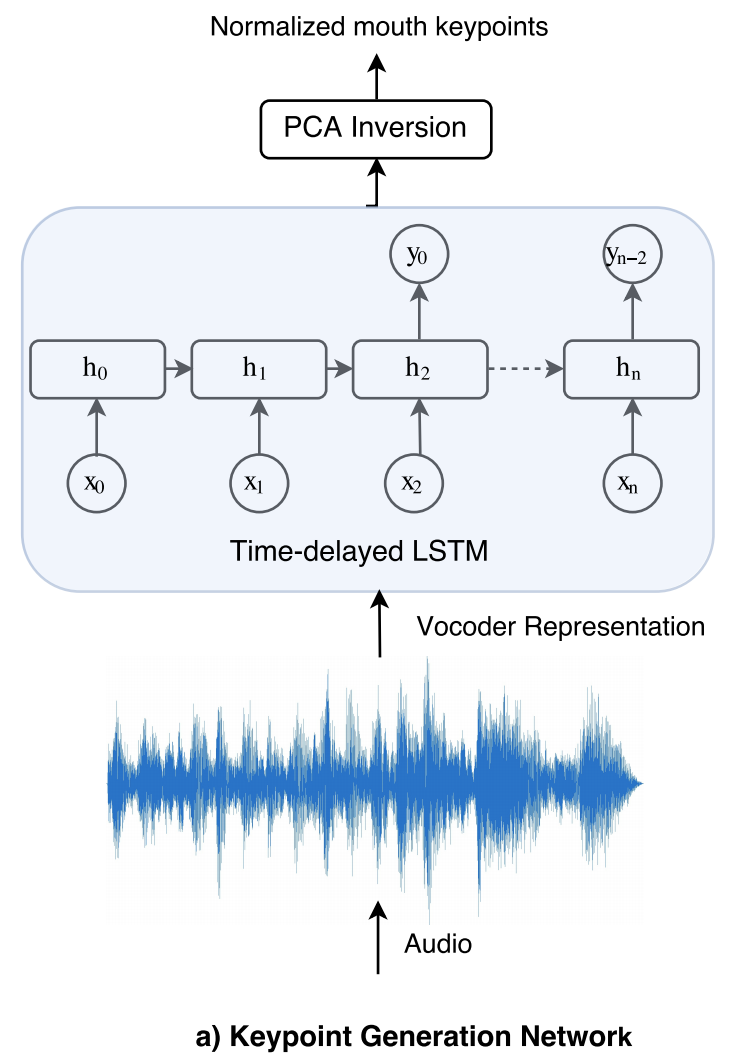}
    \end{center}
    \caption{Keypoint Generation Network}
\end{wrapfigure}

\subsection{Text-to-speech system}
We use the Char2Wav architecture \citep{sotelo2017char2wav} to generate speech using the input text. We train the speech synthesis system using the audio extracted from the videos, along with their corresponding transcripts. 
\subsection{Keypoint generation}
\begin{wrapfigure}[22]{r}{0.3\textwidth}
	\centering
    \vspace{-20pt}
    \includegraphics[scale=0.42]{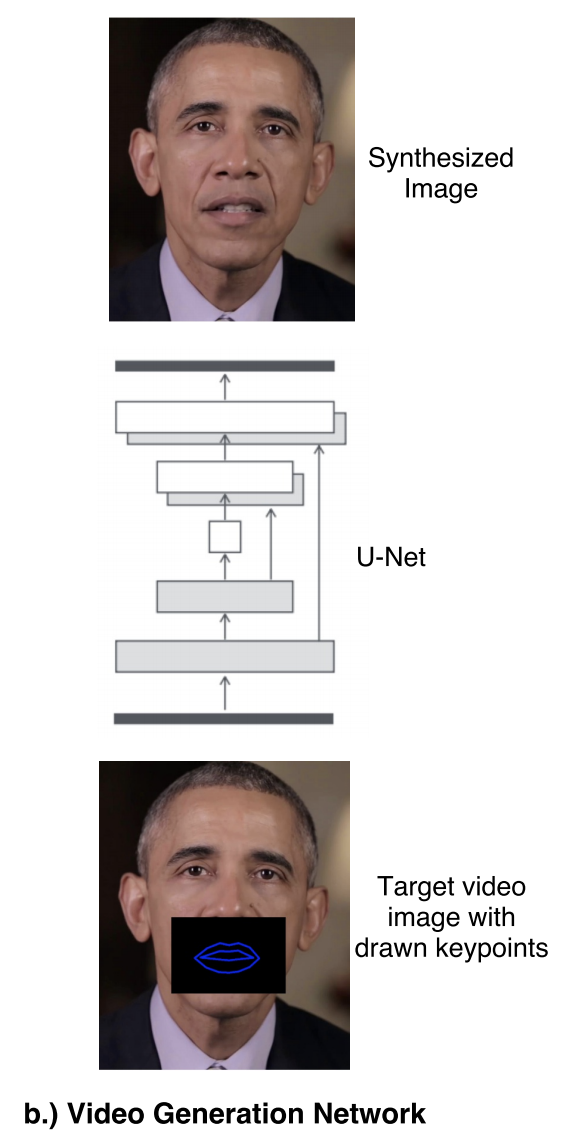}
    \caption{Video Generation Network}
\end{wrapfigure}
This module predicts the representation of the mouth shape, given the audio as input. We use spectral features to represent the audio. To compute the mouth-shape representation, we use mouth keypoints extracted from the face, and normalize the points to be invariant to image size, face location, face rotation and face size.
Normalization is crucial in the pipeline, as it makes the key-point generation compatible with any target video. We then apply PCA over the normalized mouth key-points to reduce the dimension and to decorrelate the features. We only use the most prominent principal components as the representation for mouth shape.

Regarding the network architecture, we adopt the same as \citet{suwajanakorn2017synthesizing}: we use an LSTM network (\cite{hochreiter1997long} with time-delay to predict the mouth shape representation given the audio features as input. 
\subsection{Video generation}
Our motivation behind the choice of method to perform video generation is the recent success of pix2pix (\cite{isola2016image}) as a general-purpose solution for image-to-image translation problems. This task falls within our purview, as our objective here is to translate an input face image with cropped mouth area, to an output image with in-painted mouth area, conditioned on the mouth shape representation.

To avoid explicit conditioning of mouth shape representation in the U-Net architecture, we implicitly condition by drawing an outline of mouth on the input cropped image. The network learns to leverage this outline to condition the generation of the mouth in the output.

We noticed that the keypoints generated by the recurrent network are consistent across time without abrupt changes. This allowed us to perform video generation in parallel, by synthesizing each frame in the video independently across time, given the conditioning information of the mouth keypoints. We did not need any explicit mechanism to maintain temporal consistency in the generated frames of the video.

We trained this network only using L1-loss in pixel-space and found that this objective is sufficient to learn the in-painting of the mouth and doesn't require the extra GAN objective as originally proposed in pix2pix by \cite{isola2016image}.

\newpage
\section{Supplementary Material}
\subsection{Dataset}
We showcase our approach on videos of ex-President Barack Obama, similar to \cite{suwajanakorn2017synthesizing}. We used 17 hours of video footage from his 300 weekly presidential addresses, which have the benefit to frame the president in a relatively controlled environment, with the subject in the center of the camera. 

\subsection{Data Processing}
\paragraph{Text to speech} We extract the audio from the videos and convert it to 16KHz. We extract vocoder frames from the audio using the WORLD vocoder, and use the transcript associated with the video to train the text-to-speech system.

\begin{wrapfigure}[12]{l}{0.4\textwidth}
    \centering
    \vspace{-10pt}
    \includegraphics[scale=0.4]{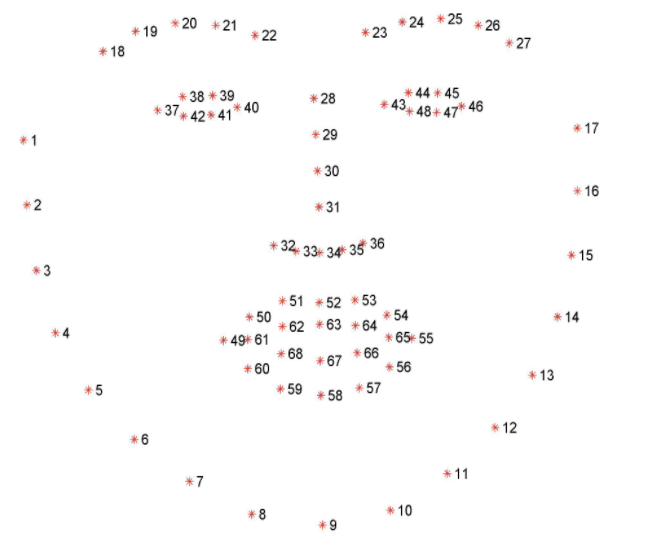}
    \caption{The 68 facial keypoints}
\end{wrapfigure}

\paragraph{Keypoint Generation} The data required for the keypoint generation component is a representation of audio for input, and a representation of mouth shape for the output.

To compute the mouth shape representation, we extract 68 facial keypoints from each frame of the video. We used the publicly available dlib facial landmark detector to detect the 68 keypoints from the image. Sample annotations performed by the detector are shown in Figure 3.

These keypoints are highly dependent on the face location, face size, in-plane and out-of-plane face rotation. These variances are due to varying zoom-levels of the camera, distance between camera and speaker, and the natural head-motion of the speaker. In an effort to remove these variances, we first mean-normalize the 68 keypoints with the center of the mouth. This converts the 68 keypoints into vectors originating from the center of the mouth, thereby making it invariant to the face location.

To remove the in-plane rotation caused due to head motion, we project the keypoints into a horizontal axis using rotation of axes. 

We make the keypoints invariant to face size, by dividing the keypoints by the norm of the 68 vectors from the center of the mouth, which serves as an approximation of face size.

Finally, we apply PCA to de-correlate the 20 normalized keypoints (40-D vector). We noticed that the first 5 PCA-coefficients capture >98\% variability in the data.

\paragraph{Video Generation} The data required for this component are image pairs, where the input face image is cropped around the mouth area and annotated with the mouth outline and the output image is the complete face. 

\begin{wrapfigure}[]{r}{0.4\textwidth}
    \centering
    \vspace{-15pt}
    \includegraphics[scale=0.38]{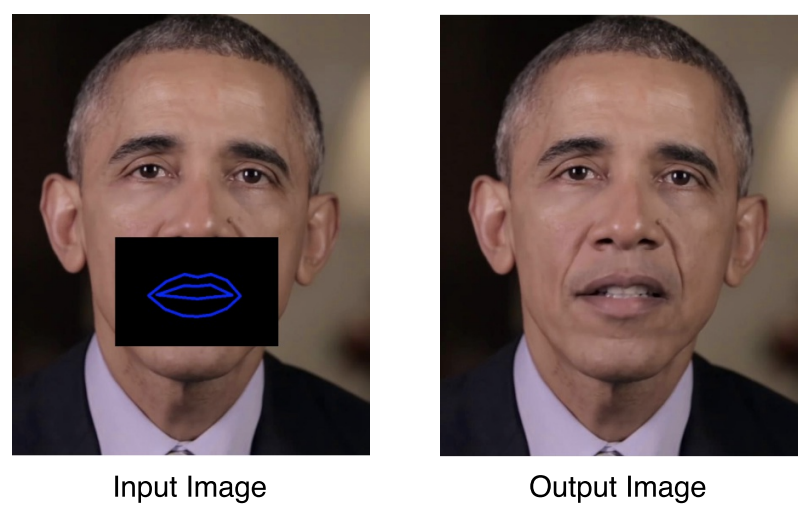}
    \caption{Sample input-output pair for the in-painting network}
    \label{fig:my_label}
\end{wrapfigure}

For this task, We extract 1 image per second of video for all 300 videos, extracting keypoints from these images using the dlib facial landmark detector. We crop the mouth area from each image using a bounding box around the mouth keypoints, and the mouth outline is drawn with keypoints 49-68 using OpenCV. Figure 4 shows a sample input / output pair.

An important aspect of the video generation process is to denormalize the generated keypoints from the previous stage of the pipeline, with the mouth location, size and rotation parameters of the target video. This ensures that the rendered mouth is visually compatible with the face in the target video.

\bibliographystyle{iclr2018_conference}
\bibliography{references}
\end{document}